\documentclass[11pt,a4paper]{article}
\usepackage[hyperref]{naaclhlt2019}
\usepackage{amsmath,amssymb}
\usepackage{framed}
\usepackage{multirow}
\usepackage{textcomp}
\usepackage{times}
\usepackage{latexsym}
\usepackage{graphicx}
\usepackage{xspace}

\newcommand{\smallpar}[1]{\paragraph{#1}}

\usepackage{url}
\usepackage{textcomp}
\usepackage{calc}

\newif\ifcomment
\commenttrue

\definecolor{darkgreen}{rgb}{0,0.75,0}
\definecolor{darkmagenta}{rgb}{0.75,0,0.75}
\definecolor{darkyellow}{rgb}{0.5,0.5,0}
\definecolor{darkcyan}{rgb}{0,0.75,0.75}
\ifcomment
\newcommand\hp[1]{\textcolor{red}{[HP: #1]}}
\newcommand\rj[1]{\textcolor{blue}{[RJ: #1]}}

\else
\newcommand\hp[1]{}
\newcommand\rj[1]{}
\fi

\newcommand{\logsumexp}{\operatorname{logsumexp}}
\newcommand{\aggop}{{\mathcal{C}}}
\newcommand{\ckb}{\mbox{CKB}\xspace}
\newcommand{\sentDV}{\mbox{\textsc{SentDrugMut}}\xspace}
\newcommand{\sentDG}{\mbox{\textsc{SentDrugGene}}\xspace}
\newcommand{\sentTriple}{\mbox{\textsc{SentLevel}}\xspace}
\newcommand{\paraDV}{\mbox{\textsc{ParaDrugMut}}\xspace}
\newcommand{\paraDG}{\mbox{\textsc{ParaDrugGene}}\xspace}
\newcommand{\paraTriple}{\mbox{\textsc{ParaLevel}}\xspace}
\newcommand{\docTriple}{\mbox{\textsc{DocLevel}}\xspace}
\newcommand{\fullmodel}{\mbox{\textsc{MultiScale}}\xspace}
\newcommand\nl[1]{``\textit{#1}''}
\newcommand{\eat}[1]{\ignorespaces}

\newcommand{\CKBCore}{CKB CORE\texttrademark}

\aclfinalcopy % Uncomment this line for the final submission
\title{Document-Level $N$-ary Relation Extraction with Multiscale Representation Learning}

\author{
    Robin Jia$^1$\Thanks{ Work done as an intern at Microsoft Research.} \qquad \qquad Cliff Wong$^2$ \qquad \qquad Hoifung Poon$^2$ \\
    $^1$ Stanford University, Stanford, California, USA \\
    $^2$ Microsoft Research, Redmond, Washington, USA \\
    \texttt{robinjia@cs.stanford.edu, \{cliff.wong,hoifung\}@microsoft.com}
}

\date{}

\begin{document}
\maketitle
\begin{abstract}
Most information extraction methods focus on binary relations expressed within single sentences. 
In high-value domains, however, $n$-ary relations are of great demand (e.g., drug-gene-mutation interactions in precision oncology).
Such relations often involve entity mentions that are far apart in the document, yet existing work on cross-sentence relation extraction is generally confined to small text spans (e.g., three consecutive sentences), which severely limits recall.
In this paper, we propose a novel multiscale neural architecture for document-level $n$-ary relation extraction.
Our system combines representations learned over various text spans throughout the document and across the subrelation hierarchy.
Widening the system's purview to the entire document maximizes potential recall.
Moreover, by integrating weak signals across the document, multiscale modeling increases precision, even in the presence of noisy labels from distant supervision.
Experiments on biomedical machine reading show that our approach substantially outperforms previous $n$-ary relation extraction methods.
\end{abstract}

\section{Introduction}

Knowledge acquisition is a perennial challenge in AI. 
In high-value domains, it has acquired new urgency in recent years due to the advent of big data.
For example, the dramatic drop in genome sequencing cost has created unprecedented opportunities for tailoring cancer treatment to a tumor's genetic composition \cite{bahcall15precision}. 
Despite this potential, operationalizing personalized medicine is difficult, in part because it requires painstaking curation of precision oncology knowledge from biomedical literature. With tens of millions of papers on PubMed, and thousands more added every day,\footnote{\url{ncbi.nlm.nih.gov/pubmed}} we are sorely in need of automated methods to accelerate manual curation. 

\begin{figure}
    \small
    \begin{framed}
        \nl{We next expressed \textcolor{darkgreen}{ALK} F1174L, 
        \textcolor{darkgreen}{ALK} F1174L/L1198P, \textcolor{darkgreen}{ALK} F1174L/\textcolor{blue}{G1123S}, and \textcolor{darkgreen}{ALK} F1174L/\textcolor{blue}{G1123D} in the original SH-SY5Y cell line.}
        
        \vspace{0.1in}
        
        \qquad (\dots 15 sentences spanning 3 paragraphs \dots)
        
        \vspace{0.1in}
        
        \nl{The 2 mutations that were only found in the neuroblastoma resistance screen (\textcolor{blue}{G1123S/D}) are located in the glycine-rich loop, which is known to be crucial for ATP and ligand binding and are the first mutations described that induce resistance to TAE684, but not to \textcolor{red}{PF02341066}.}
    \end{framed}
    \caption{Two examples of \textcolor{red}{drug}-\textcolor{darkgreen}{gene}-\textcolor{blue}{mutation} relations from a biomedical journal paper. 
    The relations are expressed across multiple paragraphs, requiring document-level extraction.}
    \label{fig:text-example}
\end{figure}

Prior work in machine reading has made great strides in sentence-level binary relation extraction.
However, generalizing extraction to $n$-ary relations poses new challenges.
Higher-order relations often involve entity mentions that are far away in the document. 
Recent work on $n$-ary relation extraction has begun to explore cross-sentence extraction \cite{peng&al17,wang&poon18}, but the scope is still confined to short text spans (e.g., three consecutive sentences), even though a document may contain hundreds of sentences and tens of thousands of words.
While this already increases the yield compared to sentence-level extraction, it still misses many relations.
For example, in Figure~\ref{fig:text-example}, the drug-gene-mutation relations between $\tt PF02341066$, $\tt ALK$, $\tt G1123S(D)$ ($\tt PF02341066$ can treat cancers with mutation $\tt G1123S(D)$ in gene $\tt ALK$) can only be extracted by substantially expanding the scope.
High-value information, such as latest medical findings, might only be mentioned once in the corpus. Maximizing recall is thus of paramount importance.

\begin{figure*}
    \centering
    \includegraphics[width=0.9\textwidth]{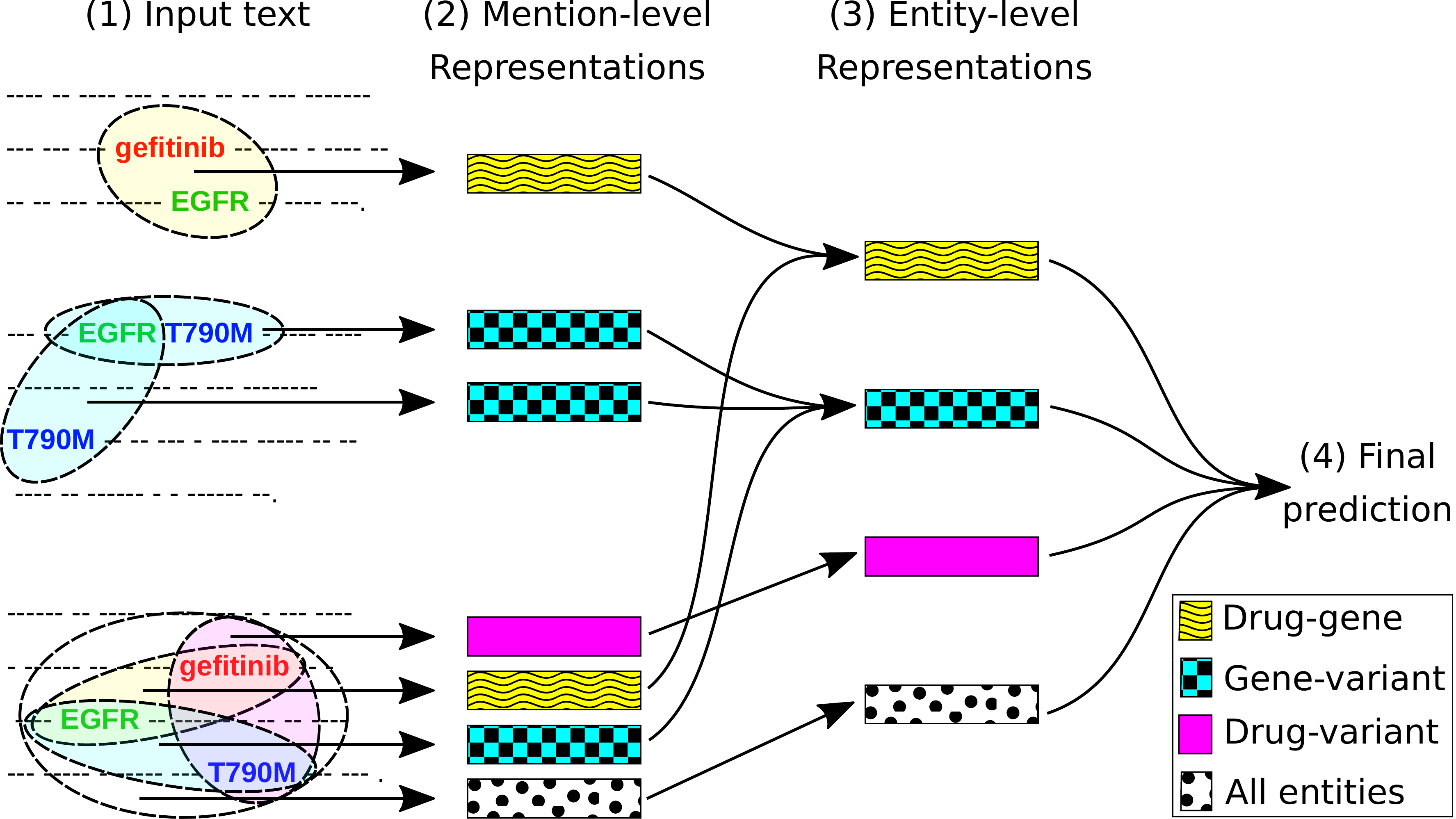}
    \caption{Multiscale representation learning for document-level $n$-ary relation extraction, an entity-centric approach that combines mention-level representations learned across text spans and subrelation hierarchy. 
    (1) Entity mentions (e.g., \textcolor{red}{gefitinib, a drug}; \textcolor{darkgreen}{EGFR, a gene}; \textcolor{blue}{T790M, a variant}) are identified from text, and mentions that co-occur within a discourse unit (e.g., paragraph) are isolated.
    (2) Within each discourse unit, mention-level representations are computed for each tuple of entity mentions. 
    These representations may correspond to the entire $n$-ary relation or subrelations over subsets of entities (\textcolor{darkmagenta}{drug-variant}, \textcolor{darkyellow}{drug-gene}, \textcolor{darkcyan}{gene-variant}).
    (3) At the document scale, mention-level representations for both the $n$-ary relation and its subrelations are combined into entity-level representations.
    (4) Entity-level representations are used to predict the relation. 
    }
    \label{fig:multiscale}
\end{figure*}

In this paper, we propose a novel multiscale neural architecture for document-level $n$-ary relation extraction.
By expanding extraction scope to the entire document, rather than restricting relation candidates to co-occurring entities in a short text span, we ensure maximum potential recall. 
To combat the ensuing difficulties in document-level extraction, such as low precision, we introduce multiscale learning,
which combines representations learned over text spans of varying scales and for various subrelations (Figure~\ref{fig:multiscale}). 
This approach deviates from past methods in several key regards. 

First, we adopt an entity-centric formulation by making a single prediction for each entity tuple occurring in a document. 
Previous $n$-ary relation extraction methods typically classify individual mention tuples, but this approach scales poorly to whole documents.
Since each entity can be mentioned many times in the same document,
applying mention-level methods leads to
a combinatorial explosion of mention tuples.
This creates not only computational challenges but also learning challenges,
as the vast majority of these tuples do not express the relation.
Our entity-centric formulation alleviates both of these problems.

Second, for each candidate tuple, prior methods typically take as input the contiguous text span encompassing the mentions.
For document-level extraction, the resulting text span could become untenably large, even though most of it is unrelated to the relation of interest. Instead, we allow discontiguous input formed by multiple discourse units (e.g., sentence or paragraph) containing the given entity mentions.

Finally, while an $n$-ary relation might not reside within a discourse unit, its subrelations might. In Figure 1, the paper first mentions a gene-mutation subrelation, then discusses a drug-mutation subrelation in a later paragraph. 
By including subrelations in our modeling, we can predict
$n$-ary relations even when all $n$ entities never co-occur in the same discourse unit.

With multiscale learning, we turn the document view from a challenge into an advantage by combining weak signals across text spans and subrelations.
Following recent work in cross-sentence relation extraction, we conduct thorough evaluation in biomedical machine reading.
Our approach substantially outperforms prior $n$-ary relation extraction methods, attaining state-of-the-art results on a large benchmark dataset recently released by a major cancer center.
Ablation studies show that multiscale modeling is the key to these gains.\footnote{
Our code and data will be available at \url{hanover.azurewebsites.net}
}
\section{Document-Level $N$-ary Relation Extraction}

Prior work on relation extraction typically formulates it as a mention-level classification problem.
Let $e_1,\dotsc,e_n$ be entity mentions that co-occur in a text span $T$. 
Relation extraction amounts to classifying whether a relation $R$ holds for $e_1,\dotsc,e_n$ in $T$.
For the well-studied case of binary relations within single sentences, $n=2$ and $T$ is a sentence.

In high-value domains, however, there is increasing demand for document-level $n$-ary relation extraction, where $n>2$ and $T$ is a full document that may contain hundreds of sentences.
For example, a molecular tumor board needs to know if a drug is relevant for treating cancer patients with a certain mutation in a given gene.
We can help the tumor board by extracting such ternary interactions from biomedical articles.
The mention-centric view of relation extraction does not scale well to this general setting.
Each of the $n$ entities may be mentioned many times in a document, resulting in a large number of candidate mention tuples, even though the vast majority of them are irrelevant to the extraction task.

In this paper, we adopt an \emph{entity-centric} formulation for document-level $n$-ary relation extraction.
We use upper case for entities ($E_1,\cdots,E_n$) and lower case for mentions ($e_1,\cdots,e_n$). 
We define an $n$-ary relation candidate to be an $(n+1)$-tuple $(E_1, \dotsc, E_n, T)$, where each entity $E_i$ is mentioned at least once in the text span $T$.
The relation extraction model is given a candidate $(E_1, \dotsc, E_n, T)$
and outputs whether or not the tuple expresses the relation $R$.\footnote{
It is easy to extend our approach to situations where $k$ mutually exclusive relations
$R_1, \dotsc, R_k$ must be distinguished, resulting in a $(k+1)$-way classification problem.}
Deciding what information to use from the various entity mentions within $T$ is now a modeling question, which we address in the next section.
\section{Our Approach: Multiscale Representation Learning}

We present a general framework for document-level $n$-ary relation extraction using multiscale representation learning.
Given a document with text $T$ and entities $E_1, \dotsc, E_n$,
we first build mention-level representations for groups of these entities
whenever they co-occur within the same discourse unit.
We then aggregate these representations across the whole document,
yielding entity-level representations for each subset of entities.
Finally, we predict whether $E_1, \dotsc, E_n$ participate in the relation
based on the concatenation of these entity-level representations.
These steps are depicted in Figure~\ref{fig:multiscale}.

\subsection{Mention-level Representation}
Let the full document $T$ be composed of discourse units $T_1, \dotsc, T_m$
(e.g., different paragraphs).
Let $T_j$ be one such discourse unit,
and suppose $e_1,\dotsc,e_n$ are entity mentions of $E_1, \dotsc, E_n$ that co-occur in $T_j$. We construct a contextualized representation for mention tuple $(e_1,\dotsc,e_n)$ in $T_j$. In this paper, we use a standard approach by applying a bi-directional LSTM (BiLSTM) to $T_j$, concatenating the hidden states for each mention, and feeding this through a single-layer neural network.
We denote the resulting vector as $r(R, e_1,\dotsc,e_n, T_j)$ for the relation $R$.

\subsection{Entity-level Representation}

Let $M(R, E_1, \dotsc, E_n, T)$ denote the set of all mention tuples 
$(e_1, \dotsc, e_n)$ and discourse units $T_j$ within $T$ such that
each $e_i$ appears in $T_j$.
We can create an entity-level representation $r(R, E_1,\dotsc,E_n, T)$ of the $n$ entities
by combining mention-level representations using an aggregation operator $\aggop$:
\[
\underset{(e_1,\dotsc,e_n, T_j)\in M(R, E_1, \dotsc, E_n, T)}{\aggop}
r(R, e_1,\dotsc,e_n,T_j)
\]

A standard choice for $\aggop$ is max pooling, which works well if it is pretty clear-cut whether a mention tuple expresses a relation. In practice, however, the mention tuples could be ambiguous and less than certain individually, yet collectively express a relation in the document. This motivates us to experiment with $\logsumexp$, the smooth version of max, where 
\[
\logsumexp(x_1,\dotsc,x_k)=\log\sum_{i=1}^k\exp(x_i).
\]
This facilitates accumulating weak signals from individual mention tuples, and our experiments show that it substantially improves extraction accuracy compared to max pooling.

\subsection{Subrelations}

For higher-order relations (i.e., larger $n$), it is less likely that they will be completely contained within a discourse unit. Often, the relation can be decomposed into subrelations over subsets of entities, each of which is more likely to be expressed in a single discourse unit.  
This motivates us to construct entity-level representations for subrelations as well.
The process is straightforward. 
Let $R_S$ be the $|S|$-ary subrelation over entities $E_{S_1},\cdots,E_{S_{|S|}}$, where $S \subseteq \{1, \dotsc, n\}$ and $|S|$ denotes its size.
We first construct mention-level representations $r(R_S, e_{S_1},\cdots, e_{S_{|S|}}, T)$ for $R_S$ and its relevant entity mentions, then combine them into an entity-level representation $r(R_S,E_{S_1},\cdots,E_{S_{|S|}},D)$ using the chosen aggregation operator $\aggop$.
We do this for every $S \subseteq \{1, \dotsc, n\}$ with $|S| \ge 2$
(including the whole set, which corresponds to the full relation $R$).
This gives us an entity-level representation for each subrelation of arity at least $2$, or equivalently, each subset of entities of size at least $2$.

\subsection{Relation Prediction}
To make a final prediction, we first concatenate all of the entity-level representations $r(R_S, E_{S_1}, \dotsc, E_{S_{|S|}}, D)$ for all $S \subseteq \{1, \dotsc, n\}$ with $|S| \ge 2$.
The concatenated representation is fed through a two-layer feedforward neural network followed by a softmax function to predict the relation type.

It is possible that for some subrelations $R_S$, all $|S|$
entities do not co-occur in any discourse unit.
When this happens, we set $r(R_S, E_{S_1}, \dotsc, E_{S_{|S|}})$
to a bias vector which is learned separately for each $R_S$.
This ensures that the concatenation is done over a fixed number of vectors, e.g., $4$ for a tenary relation (three binary subrelations and the main relation).
Importantly, this strategy enables us to make meaningful predictions for relation candidates
even if all $n$ entities never co-occur in the same discourse unit;
such candidates would never be generated by a system that only looks at single discourse units in isolation.

\subsection{Document Model}

\newcommand{\ensemble}{{\mathcal{P}}}

Our document model is actually a family of representation learning methods, conditioned on the choice of discourse units, subrelations, and aggregation operators.
In this paper, we consider sentences and paragraphs as possible discourse units.
We explore $\max$ and $\logsumexp$ as aggregation operators.
Moreover, we explore ensemble prediction as an additional aggregation method. 
Specifically, we learn a restricted multiscale model by limiting the text span to a single discourse unit (e.g., a paragraph); the model still combines representations across mentions and subrelations. 
At test time, given a full document with $m$ discourse units, we obtain independent predictions $p_1, \dotsc, p_m$ for each discourse unit. We then combine these probabilities using an ensemble operator $\ensemble$.
A natural choice for $\ensemble$ is max, though we also experiment with noisy-or:
\[
\ensemble(p_1,\cdots,p_k) = 1-\prod_{i=1}^k~(1-p_i).
\]
It is also possible to ensemble multiple models that operate on different
discourse units, using this same operator.

Our model can be trained using standard supervised or indirectly supervised methods. In this paper, we focus on distant supervision, as it is a particularly potent learning paradigm for high-value domains. Our entity-centric formulation is particularly well aligned with distant supervision, as distant supervision at the entity level is significantly less noisy compared to the mention level, so we don't need to deploy sophisticated denoising strategies such as multi-instance learning \cite{hoffmann2011knowledge}. 
\section{Experiments}
\begin{table}[t]
  \centering
  \small
  \begin{tabular}{|l|ccc|}
    \hline
    & Sentence & Paragraph & Document \\
    & level & level & level \\
    \hline
    Text Units & $2,326$ & $3,687$ & $3,362$ \\
    \hline
    Pos. Examples & 2,222 & 4,906 & 8,514 \\
    \hline
    Neg. Examples & $2,849$ & $13,371$ & $323,584$ \\
    \hline
  \end{tabular}
  \caption{Statistics of our training corpus using PMC-OA articles and distant supervision from CIVIC, GDKD, and OncoKB. 
  ``Text Units'' refers to the number of distinct sentences, paragraphs, and documents that contain a candidate triple of drug, gene, mutation.
  }
  \label{tab:pmc}
\end{table}
\begin{table}[t]
  \centering
  \small
  \begin{tabular}{|l|cc|}
    \hline
    & Development & Test \\
    \hline
    Documents & $118$ & $225$ \\
    Annotated facts & $701$ & $1,324$ \\
    \hline
    Paragraphs per document & $101$ & $105$ \\
    Sentences per document & $314$ & $320$ \\
    Words per document & $6,871$ & $7,010$ \\
    \hline
  \end{tabular}
  \caption{Statistics of the \ckb evaluation corpus.
  }
  \label{tab:ckb}
\end{table}
\subsection{Biomedical Machine Reading}
We validate our approach on a standard biomedical machine reading task: extracting drug-gene-mutation interactions from biomedical articles \cite{peng&al17,wang&poon18}. 
We cast this task as binary classification: given a drug, gene, mutation, and document in which they are mentioned, determine whether the document asserts that the mutation in the gene affects response to the drug.
For training, we use documents from the PubMed Central Open Access Subset (PMC-OA)\footnote{\url{www.ncbi.nlm.nih.gov/pmc}}.
For distant supervision, we use three existing knowledgebases (KBs) with hand-curated drug-gene-mutation facts: CIVIC,\footnote{\url{civic.genome.wustl.edu}} GDKD \cite{dienstmann&al15}, and OncoKB \cite{chakravarty&al17}.
Table~\ref{tab:pmc} shows basic statistics of this training data. 
Past methods using distant supervision often need to up-weight positive examples, due to the large proportion of negative candidates. Interestingly, we found that our document model was robust to this imbalance, as re-weighting had little effect and we didn't use it in our final results. 

Evaluating distant supervision methods is challenging, as there is often no gold-standard test set, especially at the mention level.
Prior work thus resorts to reporting sample precision (estimated proportion of correct system extractions) and absolute recall (estimated number of correct system extractions). This requires subsampling extraction results and manually annotating them. Subsampling variance also introduces noise in the estimate.

\begin{table*}[ht]
  \centering
  \small
  \begin{tabular}{|l|c|c|ccc|}
    \hline
    System & AUC & Max Recall & Precision & Recall & F1 \\
    \hline
    \textbf{Base versions} & & & & & \\
    DPL         & $24.4$ & $53.8$ & $27.3$ & $42.3$ & $33.2$ \\
    \sentTriple & $22.4$ & $36.6$ & $39.3$ & $34.7$ & $36.9$ \\
    \paraTriple & $33.1$ & $58.9$ & $36.5$ & $44.6$ & $40.1$ \\
    \docTriple  & $36.7$ & $\bf 79.0$ & $45.4$ & $38.5$ & $41.7$ \\
    \fullmodel  & $\bf 37.3$ & $\bf 79.0$ & $41.8$ & $43.3$ & $\bf 42.5$ \\
    \hline
    \textbf{+ Noisy-Or} & & & & & \\
    DPL         & $31.5$ & $53.8$ & $33.3$ & $41.5$ & $36.9$ \\
    \sentTriple & $25.3$ & $36.6$ & $39.3$ & $35.3$ & $37.2$ \\
    \paraTriple & $35.6$ & $58.9$ & $44.3$ & $40.6$ & $42.4$ \\
    \docTriple  & $36.7$ & $\bf 79.0$ & $45.4$ & $38.5$ & $41.7$ \\
    \fullmodel  & $\bf 39.7$ & $\bf 79.0$ & $48.1$ & $38.9$ & $\bf 43.0$ \\
    \hline
    \textbf{+ Noisy-Or + Gene-mutation filter} & & & & & \\
    DPL         & $39.1$ & $52.6$ & $50.5$ & $47.8$ & $49.1$ \\
    \sentTriple & $29.0$ & $35.5$ & $63.3$ & $34.2$ & $44.4$ \\
    \paraTriple & $42.1$ & $57.2$ & $50.6$ & $50.7$ & $50.7$ \\
    \docTriple  & $42.9$ & $\bf 74.4$ & $49.3$ & $46.6$ & $47.9$ \\
    \fullmodel  & $\bf 47.5$ & $\bf 74.4$ & $52.6$ & $53.0$ & $\bf 52.8$ \\
    \hline
  \end{tabular}
  \caption{Comparison of our multiscale system with restricted variants and DPL \cite{wang&poon18} on \ckb.\footnotemark
  }
  \label{tab:results}
\end{table*}
\footnotetext{corrections in supplementary section \ref{sec:supplemental_corrections}}
Instead, we used \CKBCore, a public subset of the Clinical Knowledgebase (CKB)\footnote{\url{ckbhome.jax.org}} \cite{patterson&al16}, as our gold-standard test set.
\CKBCore\ contains document-level annotation of drug-gene-mutation interactions manually curated by The Jackson Laboratory (JAX), an NCI-designated cancer center.
It is a high-quality KB containing facts from a few hundred PubMed articles for 86 genes, with minimal overlap with the three KBs we used for distant supervision.
To avoid contamination, we removed \ckb entries whose documents were used in our training data,
and split the rest into a development and test set.
See Table~\ref{tab:ckb} for statistics. 
We tuned hyperparameters and thresholds on the development set, and report results on the test set.

\subsection{Implementation Details}

We conducted standard preprocessing and entity linking, similar to \citet{wang&poon18} (see  Section~\ref{sec:preprocessing}).
Following standard practice, we masked all entities of the same type with a dummy token, to prevent the classifier from simply memorizing the facts in distant supervision.
\citet{wang&poon18} observed that many errors stemmed from incorrect gene-mutation association.
We therefore developed a simple rule-based system that predicts which gene-mutation
pairs are valid (see Section~\ref{sec:gene-mutation}).
We removed candidates that contained a gene-mutation pair that was not predicted
by the rule-based system.

\subsection{Main Results}

We evaluate primarily on area under the precision recall curve (AUC).\footnote{
We compute area using average precision, which is similar to a right Riemann sum.
This avoids errors introduced by the trapezoidal rule, which may overestimate area.}
We also report maximum recall, which is the fraction of true facts for which a candidate was generated.
Finally, we report precision, recall, and F1, using a threshold tuned to maximize F1 on the \ckb development set.

We compared our multiscale system (\fullmodel) with three restricted variants (\sentTriple, \paraTriple, \docTriple).
\sentTriple and \paraTriple restricted training and prediction to single discourse units (i.e., sentences and paragraphs), and produced a document-level prediction by applying the ensemble operator over individual discourse units.
\docTriple takes the whole document as input, with each paragraph as a discourse unit.
\fullmodel further combined \sentTriple, \paraTriple, and \docTriple using the ensemble operator.
For additional details about the models, see Section~\ref{sec:hyperparameters}.
We also compared \fullmodel with DPL \cite{wang&poon18}, the prior state of the art in cross-sentence $n$-ary relation extraction. 
DPL classifies drug-gene-mutation interactions within three consecutive sentences
using the same model architecture as \citet{peng&al17},
but incorporates additional indirect supervision such as data programming and joint inference.
We used the DPL code from the authors and produced a document-level prediction similarly using the ensemble operator.
In the base version, we used max as the ensemble operator. 
We also evaluated the effect when we used noisy-or as the ensemble operator, as well as when we applied the gene-mutation filter during postprocessing.

Table~\ref{tab:results} shows the results on the \ckb test set.
In all scenarios, our full model (\fullmodel) substantially outperforms the prior state-of-the-art system (DPL).
For example, in the best setting, using both noisy-or and the gene-mutation filter,
the full model improves over DPL by $8.4$ AUC points.
Multiscale learning is the key to this performance gain, with \fullmodel substantially outperforming more restricted variants.
Not surprisingly, expanding extraction scope from sentences to paragraphs resulted in the biggest gain, already surpassing DPL.
Conducting end-to-end learning over a document-level representation, as in \docTriple, is beneficial compared to ensembling over predictions for individual discourse units (\sentTriple, \paraTriple), especially in the base version.
Interestingly, \fullmodel still attained significant gain over \docTriple with an ensemble over \sentTriple and \paraTriple, suggesting that the document-level representation can still be improved.
In addition to prediction accuracy, the document-level models also have much more room to grow, as maximum recall is about 20 absolute points higher in \fullmodel and \docTriple, compared to \paraTriple or DPL.\footnote{The difference in actual recall is less pronounced, as we chose thresholds to maximize F1 score. We expect actual recall to increase significantly as document-level models improve, whereas the other models are closer to their ceiling.}

The ensemble operator had a surprisingly large effect, as shown by the gain when it was changed from max (base version) to noisy-or. 
This suggests that combining weak signals across multiple scales can be quite beneficial.
Our hand-crafted gene-mutation filter also improved all systems substantially,
corroborating the analysis of \newcite{wang&poon18}.
In particular, without the filter, it is hard for the document-level models to achieve high precision,
so they sacrifice a lot of recall to get good F1 scores. Using the filter helps them attain significantly higher recall while maintaining respectable precision.

\begin{figure}[t]
    \centering
    \includegraphics[width=\linewidth]{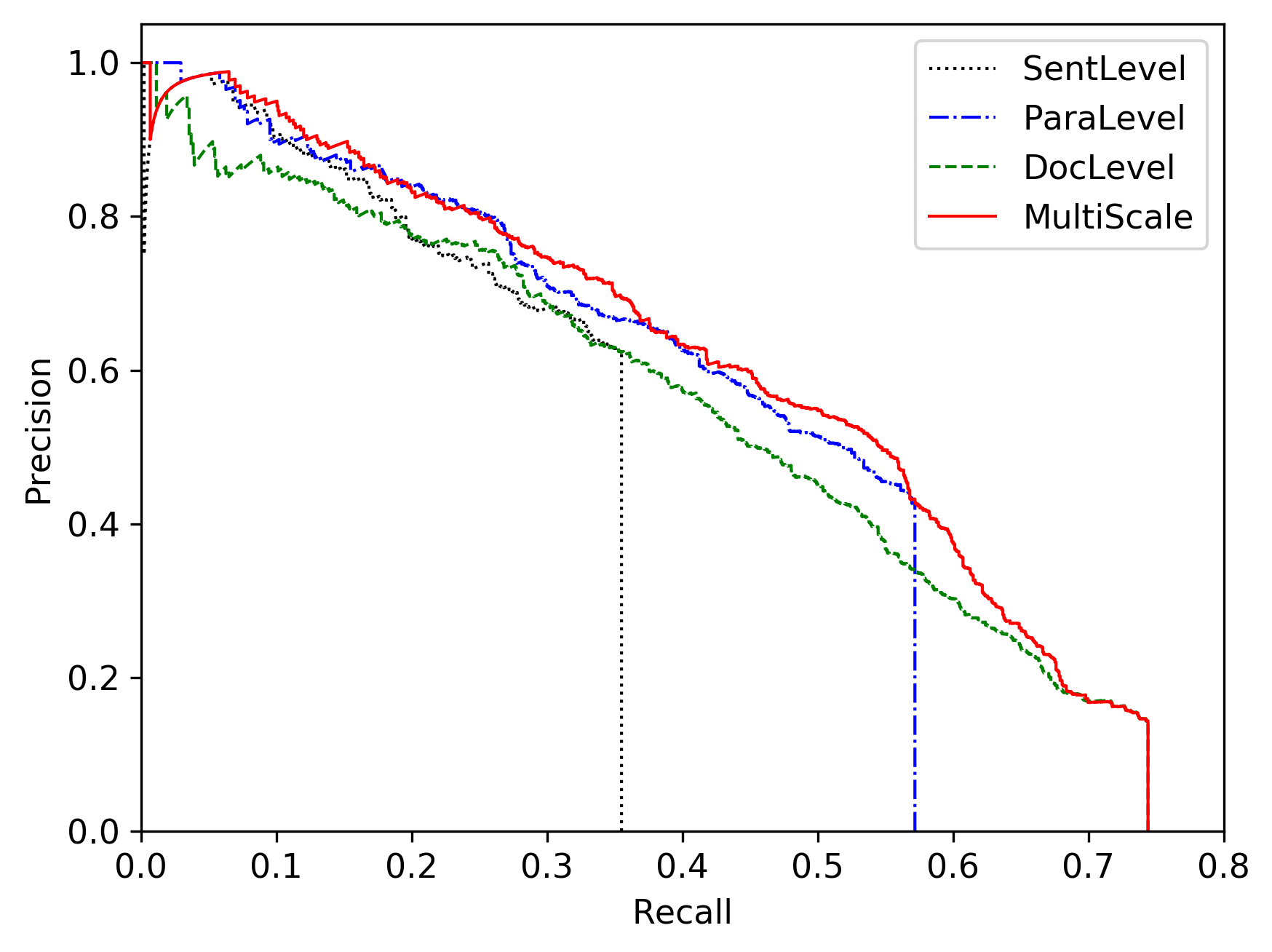}
    \caption{Precision-recall curves on \ckb (with noisy-or and gene-mutation filter).
    \fullmodel attained generally better precision than \paraTriple, and higher maximum recall like \docTriple.\footnotemark[6]
    }
    \label{fig:pr}
\end{figure}
\begin{table}[t]
  \centering
  \small
  \begin{tabular}{|l|c|c|ccc|}
    \hline
    System & AUC & MR & P & R & F1 \\
    \hline
    \fullmodel  & $\bf 47.5$ & $74.4$ & $52.6$ & $53.0$ & $\bf 52.8$ \\
    -- \sentTriple       & $47.0$ & $74.4$ & $43.0$ & $55.8$ & $48.6$ \\
    -- \paraTriple       & $45.9$ & $74.4$ & $48.8$ & $49.8$ & $49.3$ \\
    -- \docTriple        & $42.4$ & $57.2$ & $59.6$ & $44.4$ & $50.9$ \\
    \hline
  \end{tabular}
  \caption{Results on \ckb when removing either \sentTriple, \paraTriple, or \docTriple
  from the ensemble computed by \fullmodel.
  MR=max recall, P=precision, R=recall.\footnotemark[6]}
  \label{tab:ablate-models}
\end{table}
\begin{table}[t]
  \centering
  \small
  \begin{tabular}{|l|c|ccc|}
    \hline
    System & AUC & P & R & F1 \\
    \hline
    \sentTriple      & $28.3$ & $62.7$ & $35.1$ & $45.0$ \\
    \paraTriple      & $38.1$ & $47.4$ & $52.2$ & $49.7$ \\
    \docTriple       & $41.1$ & $48.2$ & $45.6$ & $46.9$ \\
    \fullmodel       & $43.7$ & $45.7$ & $51.2$ & $48.3$ \\
    \hline
  \end{tabular}
  \caption{Results on CKB after replacing $\tt logsumexp$ with $\max$ (with noisy-or and gene-mutation filter). P=precision, R=recall. Max recall same as in Table~\ref{tab:results}.\footnotemark[6]}
   \label{tab:logsumexp}
\end{table}

Figure~\ref{fig:pr} shows the precision-recall curves for the four models (with noisy-or and gene-mutation filter).
\docTriple has higher maximum recall than \paraTriple, but generally lower precision at the same recall level.
By ensembling all three variants, \fullmodel achieves the best combination: 
it generally improves precision while capturing more cross-paragraph relations.
This can also be seen in Table~\ref{tab:ablate-models}, where we 
ablate each of the three variants used by \fullmodel. 
All three variants in the ensemble contributed to overall performance.

We use $\logsumexp$ as the aggregation operator to combine mention-level representations into an entity-level one.
If we replace it with max pooling, the performance drops substantially across the board, as shown in Table~\ref{tab:logsumexp}.
For example, \fullmodel lost 3.8 absolute points in AUC. 
Such difference is also observed in \citet{verga&al18}.
As in comparing ensemble operators, this demonstrates the benefit of combining weak signals using a multiscale representation.

\subsection{Cross-sentence and Cross-paragraph Extractions}
\begin{figure}
    \centering
    \includegraphics[width=\linewidth]{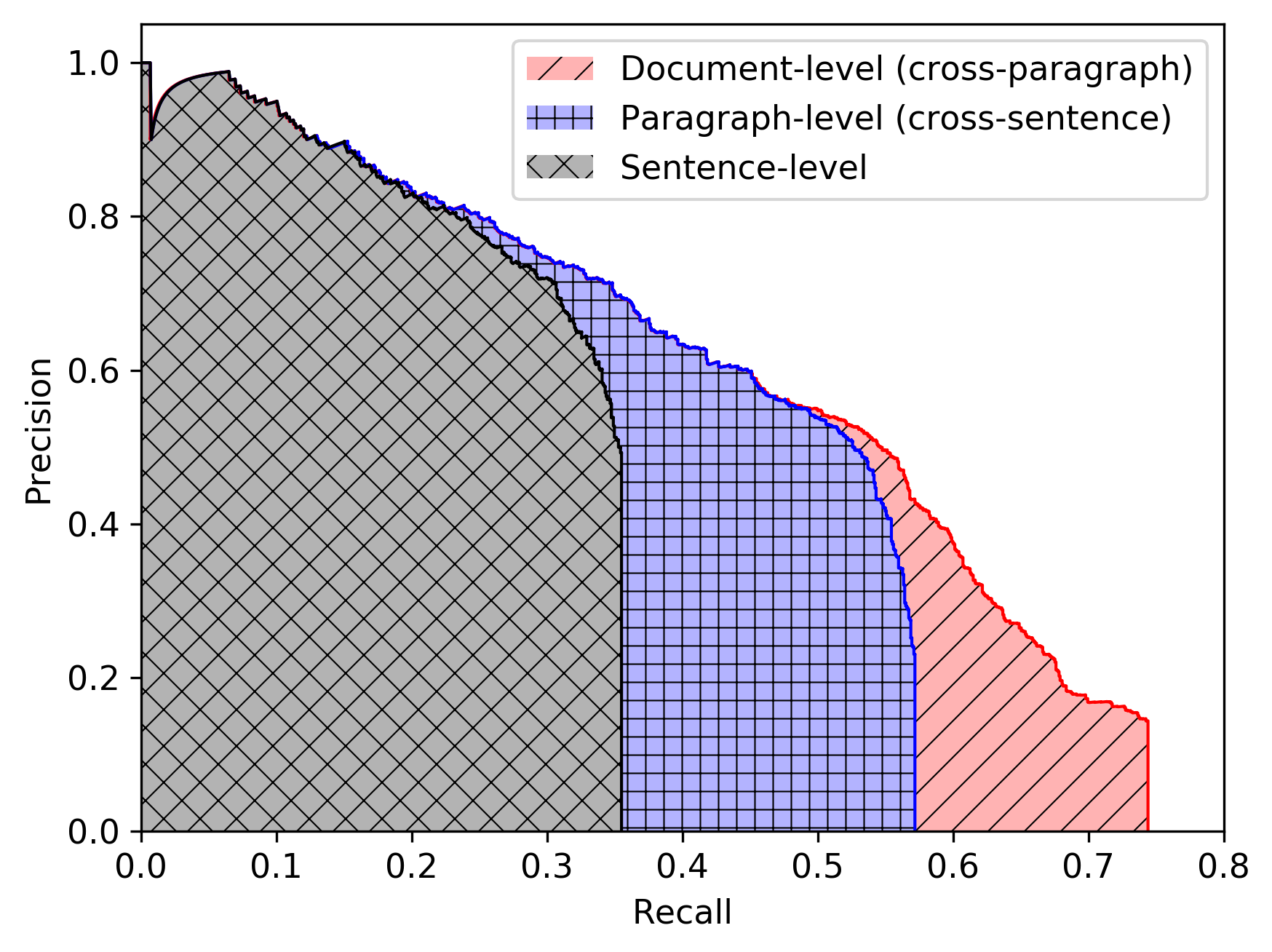}
    \caption{
    Breakdown of \fullmodel recall based on whether entities in a correctly extracted fact occurred within a single sentence, cross-sentence but within a single paragraph, or only cross-paragraph. 
    Adding cross-sentence and cross-paragraph extractions is important for high recall.
    \footnotemark[6]}
    \label{fig:cross-para}
\end{figure}

Compared to standard sentence-level extraction, our method can extract relations among entities that never co-occur in the same sentence or even paragraph.
Figure~\ref{fig:cross-para} shows the proportion of correctly predicted facts by \fullmodel that are expressed across paragraph or sentence boundaries.
\fullmodel can substantially improve the recall by making additional cross-sentence and cross-paragraph extractions.
We manually inspected twenty correct cross-paragraph extractions (with the chosen threshold for the precision/recall numbers in Table~\ref{tab:results}) and found that our model was able to handle some interesting linguistic phenomena.
Often, a paper would first describe the mutations present in a patient cohort, and later describe the effects of drug treatment.
There are also instances of bridging anaphora, for example via cell lines.
One paper first stated the gene and mutation for a cell line \nl{The \textbf{FLT3}-inhibitor resistant cells Ba/F3-ITD+691, \underline{Ba/F3-ITD+842}, \dots, which harbored FLT-ITD plus F691L, \textbf{Y842C}, \dots mutations\dots}, and later stated the drug effect on the cell line \nl{\textbf{E6201} also demonstrated strong anti-proliferative effects in \textbf{FLT3}-inhibitor resistant cells\dots such as Ba/F3-ITD+691, \underline{Ba/F3-ITD+842} \dots}.

\subsection{Subrelation Decomposition}
\begin{table}[t]
  \centering
  \small
  \begin{tabular}{|l|c|c|ccc|}
    \hline
    System & AUC & MR & P & R & F1 \\
    \hline
    \textbf{Base version} & & & & & \\
    \sentDV           & $31.0$ & $40.8$ & $60.0$ & $40.7$ & $48.5$ \\
    \sentDG           & $17.9$ & $64.2$ & $31.4$ & $27.9$ & $29.5$ \\
    \paraDV           & $39.9$ & $57.7$ & $49.3$ & $50.3$ & $49.8$ \\
    \paraDG           & $19.9$ & $68.9$ & $32.1$ & $18.9$ & $23.8$ \\
    \hline
    \textbf{+ Noisy-Or} & & & & & \\
    \sentDV           & $32.6$ & $40.8$ & $61.1$ & $39.4$ & $47.9$ \\
    \sentDG           & $23.5$ & $64.2$ & $36.3$ & $34.2$ & $35.2$ \\
    \paraDV           & $42.0$ & $57.7$ & $49.9$ & $51.5$ & $50.7$ \\
    \paraDG           & $26.1$ & $68.9$ & $46.1$ & $29.5$ & $36.0$ \\
    \hline
  \end{tabular}
  \caption{Results of subrelation decomposition baselines on \ckb, with the gene-mutation filter.
  MR=max recall, P=precision, R=recall.}
  \label{tab:subrelation}
\end{table}

As a baseline, we also consider a different document-level strategy where we decompose the $n$-ary relation into subrelations of lower arity, train independent classifiers for them, then join the subrelation predictions into one for the $n$-ary relation. 
We found that with distant supervision, the gene-mutation subrelation classifier was too noisy. Therefore, we focused on training drug-gene and drug-mutation classifiers, and joined each with the rule-based gene-mutation predictions to make ternary predictions.
Table~\ref{tab:subrelation} shows the results on \ckb. 
The paragraph-level drug-mutation model is quite competitive, which benefits from the fact that the gene-mutation associations in a document are unique. This is not true in general $n$-ary relations.
Still, it trails \fullmodel by a large margin in predictive accuracy, and with an even larger gap in the potential upside (i.e., maximum recall).
The drug-gene model has higher maximum recall, but much worse precision.
This low precision is expected, as it is usually not valid to assume that if a drug and gene interact, then all possible mutations in the gene will have an effect on the drug response.

\subsection{Error Analysis}

While much higher compared to other systems, the maximum recall for \fullmodel is still far from 100\%.
For over 20\% of the relations, we can't find all three entities in the document.
In many cases, the missing entities are in figures or supplements, beyond the scope of our extraction.
Some mutations are indirectly referenced by well-known cell lines.
There are also remaining entity linking errors (e.g., due to missing drug synonyms).

We next manually analyzed some sample prediction errors.
Among 50 false positive errors, we found a significant portion of them were actually true mentions in the paper but were excluded by curators due to additional curation criteria. For example, \ckb does not curate a fact referenced in related work, or if they deem the empirical evidence as insufficient. This suggests the need for even higher-order relation extraction to cover these aspects.
We also inspected 50 sample false negative errors.
In 40\% of the cases, the textual evidence is vague and requires corroboration from a table or figure.
In most of the remaining cases, there is direct textual evidence, though they require cross-paragraph reasoning (e.g., bridging anaphora). While \fullmodel was able to process such phenomena sometimes, there is clearly much room to improve.

\section{Related Work}

\paragraph{$N$-ary relation extraction} 
Prior work on $n$-ary relation extraction generally follows Davidsonian semantics by reducing the $n$-ary relation to $n$ binary relations between the reified relation and its arguments, a.k.a. {\em slot filling}.
For example, early work on the Message Understanding Conference (MUC) dataset aims to identify event participants in news articles
\cite{chinchor1998overview}.
More recently, there has been much work in extracting semantic roles for verbs, as in semantic role labeling \cite{palmer2010semantic}, as well as properties for popular entities, as in Wikipedia Infobox \cite{wu&weld07} and TAC KBP\footnote{\url{http://www.nist.gov/tac/2016/KBP/ColdStart/index.html}}.
In biomedicine, the BioNLP Event Extraction Shared Task aims to extract genetic events such as expression and regulation \cite{kim&al09}.
These approaches typically assume that the whole document refers to a single coherent event, or require an event anchor (e.g., verb in semantic role labeling and trigger word in event extraction). 
We instead follow recent work in cross-sentence $n$-ary relation extraction \cite{peng&al17,wang&poon18,song2018nary}, which does not have these restrictions.

\paragraph{Document-level relation extraction}
Most information extraction work focuses on modeling and prediction within sentences \cite{surdeanu&ji14}.
\newcite{duan&al17} introduces a pretrained document embedding to aid event detection, but their extraction is still at the sentence level.
Past work on cross-sentence extraction often relies on explicit coreference annotations or the assumption of a single event in the document  \cite{wick&al06,gerber&chai10,swampillai&stevenson11,yoshikawa&al11,koch&al14,yang&mitchell16}. 
Recently, there has been increasing interest in general cross-sentence relation extraction \cite{quirk&poon17,peng&al17,wang&poon18}, but their scope is still limited to short text spans of a few consecutive sentences.
These methods all extract relations at the mention level,
which does not scale to whole documents due to the combinatorial explosion of relation candidates.
\newcite{wu&al18} applies manually crafted rules to heavily filter the candidates.
We instead adopt an entity-centric approach and combine mention-level representations to create an entity-level representation for extraction.
\newcite{mintz2009distant} aggregates mention-level features into entity-level ones within a document, but they only consider binary relations within single sentences.
\citet{kilicoglu2016inferring} used hand-crafted features to improve cross-sentence extraction, 
but they focus on binary relations, and their documents are limited to abstracts, which are substantially shorter than the full-text articles we consider.
\newcite{verga&al18} applies self-attention to combine the representations of all mention pairs into an entity pair representation, which can be viewed a special case of our framework. Their work is also limited to binary relations and abstracts, rather than full documents.

\paragraph{Multiscale modeling}
Deep learning on long sequences can benefit from multiscale modeling that accounts for varying scales in the discourse structure. Prior work focuses on generative learning such as language modeling \cite{chung&al17}.
We instead apply multiscale modeling to discriminative learning for relation extraction. In addition to modeling various scales of discourse units (sentence, paragraph, document), we also combine mention-level representations into an entity-level one, as well as sub-relations of the $n$-ary relation.
\newcite{mcdonald&al05} learn $\binom{n}{2}$ pairwise relation classifiers, then construct maximal cliques of related entities, which also bears resemblance to our subrelation modeling. However, our approach incorporates the entire subrelation hierarchy, provides a principled end-to-end learning framework, and extracts relations from the whole document rather than within single sentences.

\smallpar{Distant supervision}
Distant supervision has emerged as a powerful paradigm to generate large but potentially noisy labeled datasets
\cite{craven1999constructing,mintz2009distant}.
A common denoising strategy applies multi-instance learning by treating mention-level labels as latent variables \cite{hoffmann2011knowledge}.
Noise from distant supervision increases as extraction scope expands beyond single sentences, motivating a variety of indirect supervision approaches \cite{quirk&poon17,peng&al17,wang&poon18}.
Our entity-centric representation and multiscale modeling provide an orthogonal approach to combat noise by combining weak signals spanning various text spans and subrelations.
\section{Conclusion}
We propose a multiscale, entity-centric approach for document-level $n$-ary relation extraction. 
We vastly increase maximum recall by scoring document-level candidates.
Meanwhile, we preserve precision with a multiscale approach that combines representations learned across the subrelation hierarchy and text spans of various scales.
Our method substantially outperforms prior cross-sentence $n$-ary relation extraction approaches in the high-value domain of precision oncology.

Our document-level view opens opportunities for multimodal learning by integrating information from tables and figures \cite{wu2018fonduer}.
We used the ternary drug-gene-mutation relation as a running example in this paper, but knowledge bases often store additional fields such as
effect (sensitive or resistance), cancer type (solid tumor or leukemia), and evidence (human trial or cell line experiment).
It is straightforward to apply our method to such higher-order relations.
Finally, it will be interesting to validate our approach in a real-world assisted-curation setting, where a machine reading system proposes candidate facts to be verified by human curators.

\section*{Acknowledgements}
We thank Sara Patterson and Susan Mockus for guidance on precision oncology knowledge curation and CKB data, Hai Wang for help in running experiments with deep probabilistic logic, and Tristan Naumann, Rajesh Rao, Peng Qi, John Hewitt, and the anonymous reviewers for their helpful comments.
R.J. is supported in part by an NSF Graduate Research Fellowship under Grant No. DGE-114747.

\bibliography{all}
\bibliographystyle{acl_natbib}

\appendix
\section{Appendices}
\label{sec:supplemental}

\subsection{Preprocessing}
\label{sec:preprocessing}
Full-text documents in this study were obtained from PMC.
The text was first tokenized using NLTK\footnote{https://www.nltk.org/}, then entities were extracted using a combination of regular expressions and dictionary lookups.
To identify mutation mentions, we applied a regular expression rule for missense mutations.
To identify gene mentions, we used dictionary lookup from the HUGO Gene Nomenclature Committee (HGNC)\footnote{https://www.genenames.org/} dataset.
To identify drug mentions, we used dictionary lookup from a curated list of drugs and their synonyms. 
For our training set, our list of drugs consists of all the drugs present in the distant supervision knowledge bases and selected cancer-related drugs from DrugBank\footnote{https://www.drugbank.ca/} (770 drugs total). For our test set, our drug dictionary consists of all the drugs in CKB (1119 drugs).

\subsection{Gene-mutation Rule-based System}
\label{sec:gene-mutation}
Here we describe our rule-based system for linking mutations and genes within a document.
We first generate a global mapping of mutations to sets of genes by combining publicly-available mutation-gene datasets (COSMIC\footnote{https://cancer.sanger.ac.uk/cosmic}, COSMIC Cell Lines Project\footnote{https://cancer.sanger.ac.uk/cell\textunderscore lines}, CIViC\footnote{https://civicdb.org}, and OncoKB\footnote{http://oncokb.org/}).
We then augment this mapping by finding the gene that most frequently co-occurs with each mutation in all of PubMed Central (PMC) full-text articles based on three high-precision rules:
\begin{enumerate}
    \item Gene and mutation are in the same token (e.g., "EGFR-T790M")
    \item Gene token is followed by mutation token (e.g., "EGFR T790M")
    \item Gene token is followed by a token of any single character and then followed by mutation token (e.g., "EGFR - T790M")
\end{enumerate}
For each mutation, we start with the first rule, and find all text matches for a gene with that mutation and rule.
If we found at least one match, we add the gene that occurred in the most matches to the global map. Otherwise, we repeat with the next rule.

Each mutation in the global mutation-gene map is mapped to more than $20$ genes on average. 
However, within the context of a document, each mutation is (usually) associated with just a single gene.
Given a document containing a mutation, we associate that mutation with the gene that (1) is in the global mutation-gene map for that mutation, and (2) appears closest to any mutation mention in the document.
%This technique was able to provide gene mappings for $88.7\%$ of the mutations with $96.6\%$ precision in all of CKB that we have full-text for. 

To associate genes for the remaining mutations, we apply two recall-friendly regular expression rules within that document:

\begin{enumerate}
    \setcounter{enumi}{3}
    \item Mutation is in same sentence as ``\emph{GENE} mut''
    \item Mutation is in same paragraph as ``\emph{GENE} mutation''
\end{enumerate}

We choose the first gene in the document that satisfies one of the two rules, in the above order. If there is still no matching gene at this point, the most frequent gene in the document is selected for that mutation.

\subsection{Model Details}
\label{sec:hyperparameters}
We used $200$-dimensional word vectors, initialized with word2vec vectors trained on a biomedical text corpus \citep{pyysalo&al13}.
We updated these vectors during training.
At each step, our BiLSTM received as input a concatenation of the word vectors and a $100$-dimensional embedding of the index of the current discourse unit within the document.
Following \citet{vaswani2017attention}, we used sinusoidal embeddings to represent these indices.
We used a single-layer bidirectional LSTM with a $200$-dimensional hidden state.
Mention-level representations were $400$-dimensional and computed from BiLSTM hidden states using a single linear layer followed by the $\tanh$ activation function.
For the final prediction layer, we used a two-layer feedforward network with $400$ hidden units and ReLU activation function.
We train using the Adam optimizer \citep{kingma2014adam} with learning rate of $1 \times 10^{-5}$.
During training, we consider each document to be a single batch, which allows us to reuse computation for different relation candidates in the same document.
\newpage
\section{Corrections}
\label{sec:supplemental_corrections}

The authors note that an error in the source code caused minor incorrectness in Figure 3, Figure 4, Table 3, Table 4, and Table 5. This error does not affect the conclusions of the article. The corrected figures and tables appear below.

\begin{table*}[ht]
  \centering
  \small
  \begin{tabular}{|l|c|c|ccc|}
    \hline
    System & AUC & Max Recall & Precision & Recall & F1 \\
    \hline
    \textbf{Base versions} & & & & & \\
    DPL         & $24.4$ & $53.8$ & $27.3$ & $42.3$ & $33.2$ \\
    \sentTriple & $22.4$ & $36.6$ & $38.9$ & $35.5$ & $37.1$ \\
    \paraTriple & $32.8$ & $58.9$ & $35.6$ & $44.3$ & $39.5$ \\
    \docTriple  & $\bf 37.0$ & $\bf 79.0$ & $43.3$ & $41.9$ & $\bf 42.6$ \\
    \fullmodel  & $36.9$ & $\bf 79.0$ & $38.5$ & $46.2$ & $42.0$ \\
    \hline
    \textbf{+ Noisy-Or} & & & & & \\
    DPL         & $31.5$ & $53.8$ & $33.3$ & $41.5$ & $36.9$ \\
    \sentTriple & $25.4$ & $36.6$ & $38.4$ & $35.9$ & $37.1$ \\
    \paraTriple & $35.5$ & $58.9$ & $45.5$ & $39.2$ & $42.1$ \\
    \docTriple  & $37.0$ & $\bf 79.0$ & $43.3$ & $42.0$ & $\bf 42.6$ \\
    \fullmodel  & $\bf 39.6$ & $\bf 79.0$ & $48.7$ & $37.8$ & $\bf 42.6$ \\
    \hline
    \textbf{+ Noisy-Or + Gene-mutation filter} & & & & & \\
    DPL         & $39.1$ & $52.6$ & $50.5$ & $47.8$ & $49.1$ \\
    \sentTriple & $29.0$ & $35.5$ & $63.2$ & $34.8$ & $44.9$ \\
    \paraTriple & $42.0$ & $57.2$ & $47.8$ & $53.6$ & $50.6$ \\
    \docTriple  & $43.0$ & $\bf 74.4$ & $51.1$ & $46.3$ & $48.6$ \\
    \fullmodel  & $\bf 47.3$ & $\bf 74.4$ & $54.3$ & $49.3$ & $\bf 51.7$ \\
    \hline
  \end{tabular}
  \caption*{Table 3: Comparison of our multiscale system with restricted variants and DPL \cite{wang&poon18}
  }
  \label{tab:results-corr}
\end{table*}

\begin{table*}[ht]
  \centering
  \small
  \begin{tabular}{|l|c|c|ccc|}
    \hline
    System & AUC & MR & P & R & F1 \\
    \hline
    \fullmodel  & $\bf 47.3$ & $74.4$ & $54.3$ & $49.3$ & $\bf 51.7$ \\
    -- \sentTriple       & $46.8$ & $74.4$ & $42.6$ & $56.4$ & $48.5$ \\
    -- \paraTriple       & $45.8$ & $74.4$ & $51.2$ & $48.8$ & $49.9$ \\
    -- \docTriple        & $42.4$ & $57.2$ & $59.4$ & $43.9$ & $50.5$ \\
    \hline
  \end{tabular}
  \caption*{Table 4: Results on \ckb when removing either \sentTriple, \paraTriple, or \docTriple
  from the ensemble computed by \fullmodel.
  MR=max recall, P=precision, R=recall.}
  \label{tab:ablate-models-corr}
\end{table*}

\begin{table*}[ht]
  \centering
  \small
  \begin{tabular}{|l|c|ccc|}
    \hline
    System & AUC & P & R & F1 \\
    \hline
    \sentTriple      & $28.4$ & $63.0$ & $34.8$ & $44.8$ \\
    \paraTriple      & $38.4$ & $47.4$ & $51.8$ & $49.5$ \\
    \docTriple       & $43.5$ & $56.1$ & $46.4$ & $50.8$ \\
    \fullmodel       & $43.9$ & $49.5$ & $50.7$ & $50.1$ \\
    \hline
  \end{tabular}
  \caption*{Table 5: Results on CKB after replacing $\tt logsumexp$ with $\max$ (with noisy-or and gene-mutation filter). P=precision, R=recall. Max recall same as in Table 3.}
   \label{tab:logsumexp-corr}
\end{table*}

\begin{figure}[ht]
    \centering
    \includegraphics[width=\linewidth]{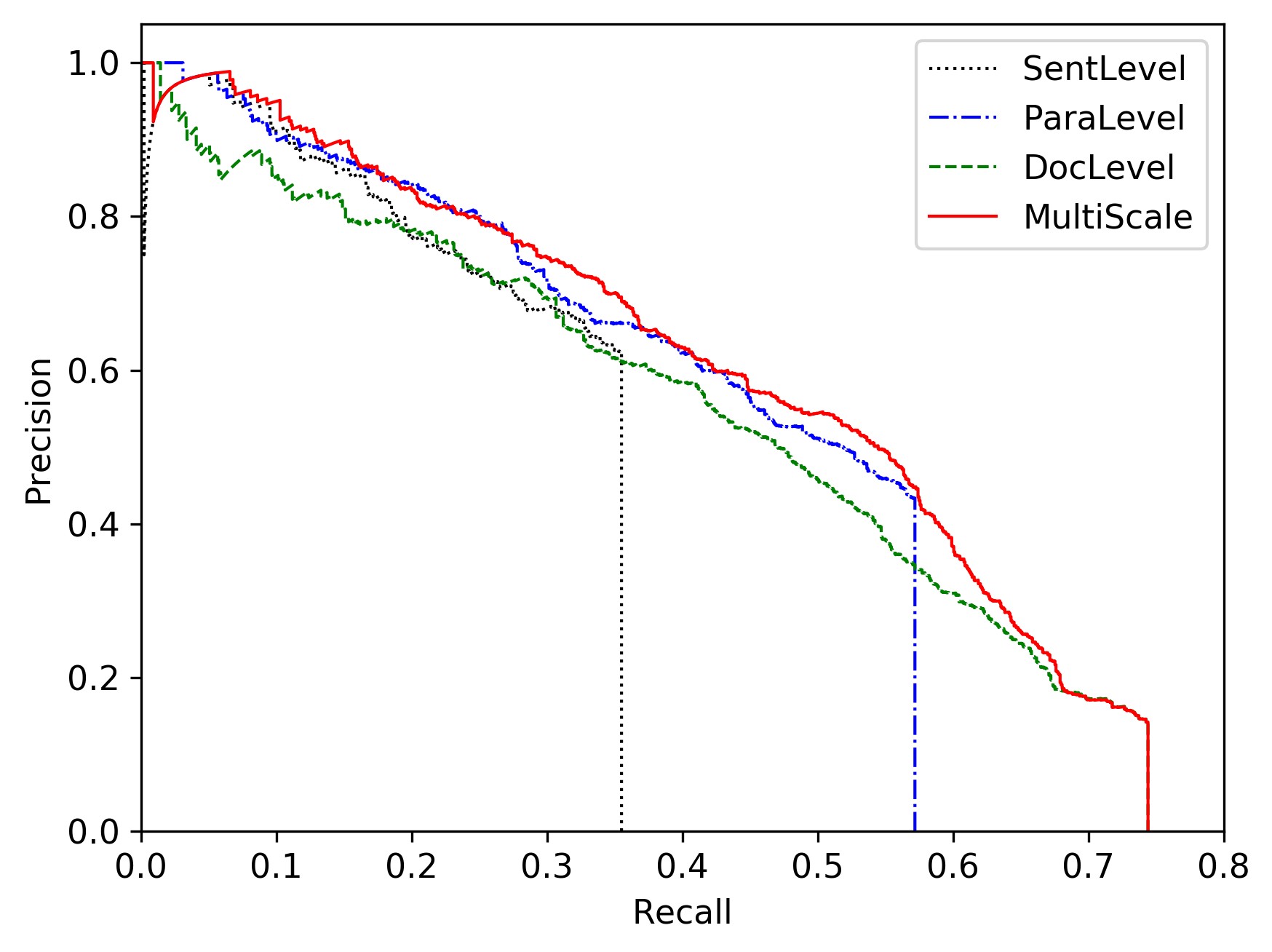}
    \caption*{Figure 3: Precision-recall curves on \ckb (with noisy-or and gene-mutation filter).
    \fullmodel attained generally better precision than \paraTriple, and higher maximum recall like \docTriple.
    }
    \label{fig:pr-corr}
\end{figure}

\begin{figure}[ht]
    \centering
    \includegraphics[width=\linewidth]{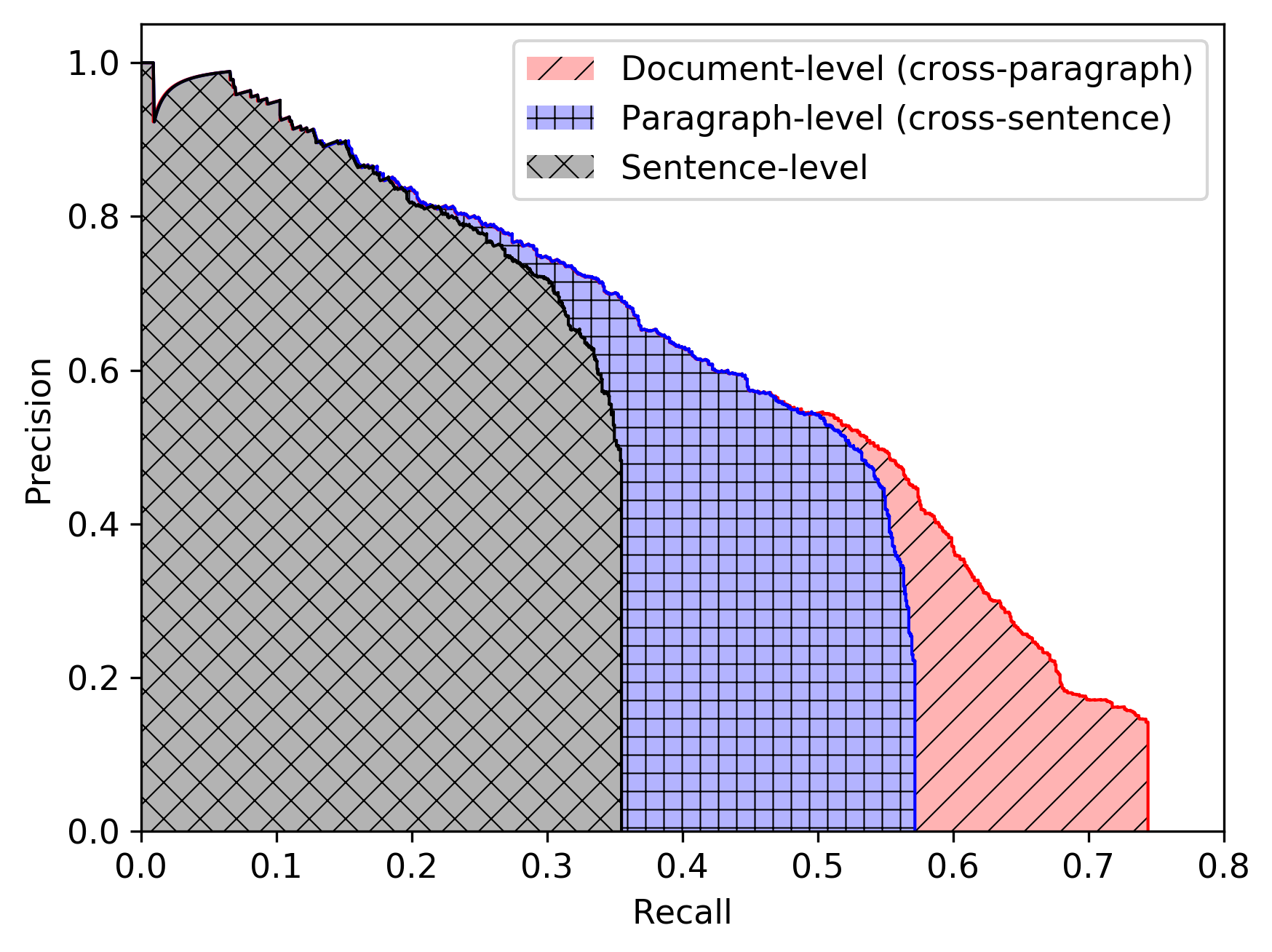}
    \caption*{
    Figure 4: Breakdown of \fullmodel recall based on whether entities in a correctly extracted fact occurred within a single sentence, cross-sentence but within a single paragraph, or only cross-paragraph. 
    Adding cross-sentence and cross-paragraph extractions is important for high recall.
    }
    \label{fig:cross-para-corr}
\end{figure}
\end{document}